\documentclass[11pt,a4paper]{article}
\usepackage{authblk}
\usepackage[hyperref]{naaclhlt2018}
\usepackage{times}
\usepackage{latexsym}

\usepackage{url}
\usepackage{booktabs}
\usepackage{graphicx}
\usepackage{array}
\usepackage{amsmath}
\usepackage{multirow}

\aclfinalcopy 
\def\aclpaperid{1070} 

\newcolumntype{C}[1]{>{\centering\let\newline\\\arraybackslash\hspace{0pt}}m{#1}}

\title{Neural Models for Reasoning over Multiple Mentions using Coreference}

\author[1]{Bhuwan Dhingra}
\author[2]{Qiao Jin}
\author[1]{Zhilin Yang}
\author[1]{\\William W. Cohen}
\author[1]{Ruslan Salakhutdinov}
\affil[1]{Carnegie Mellon University}
\affil[2]{University of Pittsburgh}
\affil[ ]{{\tt \{bdhingra,zhiliny,wcohen,rsalakhu\}@cs.cmu.edu, qij9@pitt.edu}}

\date{}

\begin{document}
\maketitle
\begin{abstract}
Many problems in NLP require aggregating information from multiple mentions of the same entity which may be far apart in the text. Existing Recurrent Neural Network (RNN) layers are biased towards short-term dependencies and hence not suited to such tasks. We present a recurrent layer which is instead biased towards coreferent dependencies. The layer uses coreference annotations extracted from an external system to connect entity mentions belonging to the same cluster. Incorporating this layer into a state-of-the-art reading comprehension model improves performance on three datasets -- Wikihop, LAMBADA and the bAbi AI tasks -- with large gains when training data is scarce.
\end{abstract}

\section{Introduction}

A long-standing goal of NLP is to build systems capable of reasoning about the information present in text. One important form of reasoning for Question Answering (QA) models is the ability to aggregate information from multiple mentions of entities. We call this \textit{coreference-based reasoning} since multiple pieces of information, which may lie across sentence, paragraph or document boundaries, are tied together with the help of referring expressions which denote the same real-world entity. Figure \ref{fig:reference-based-reasoning} shows examples.

QA models which directly read text to answer questions (commonly known as Reading Comprehension systems) \citep{hermann2015teaching,seo2016bidirectional}, typically consist of RNN layers. RNN layers have a bias towards \textit{sequential recency} \citep{dyer2017linguistic}, i.e. a tendency to favor short-term dependencies. Attention mechanisms alleviate part of the issue, but empirical studies suggest RNNs with attention also have difficulty modeling long-term dependencies \citep{daniluk2017frustratingly}. We conjecture that when training data is scarce, and inductive biases play an important role, RNN-based models would have trouble with coreference-based reasoning.

\begin{figure}[!htbp]
\small
\centering
\begin{tabular}{| p{7cm} |} 
\hline
\emph{Context:} [...] \textbf{mary} got the football there [...] \textbf{mary} went to the \underline{bedroom} [...] \textbf{mary} travelled to the hallway [...]\\
\emph{Question:} where was the football before the hallway ?
\vspace{0.03in}
\end{tabular}

\begin{tabular}{| p{7cm} |} 
\hline
\vspace{0.01in}
\emph{Context:}  Louis-Philippe Fiset [...] was a local physician and politician in the \textbf{Mauricie} area [...] is located in the \textbf{Mauricie} region of Quebec, \underline{Canada} [...]\\
\emph{Question:} country of citizenship -- louis-philippe fiset ?\\
\hline
\end{tabular}
\caption{\small Example questions which require coreference-based reasoning from the bAbi dataset (top) and Wikihop dataset (bottom). Coreferences are in bold, and the correct answers are underlined.}
\vspace{0.02in}
\label{fig:reference-based-reasoning}
\end{figure}


At the same time, systems for coreference resolution have seen a gradual increase in accuracy over the years \citep{durrett2013easy,wiseman2016learning,lee2017end}. Hence, in this work we use the annotations produced by such systems to adapt a standard RNN layer by introducing a bias towards \textit{coreferent recency}. Specifically, given an input sequence and coreference clusters extracted from an external system, we introduce a term in the update equations for Gated Recurrent Units (GRU) \citep{cho2014learning} which depends on the hidden state of the coreferent antecedent of the current token (if it exists).
This way hidden states are propagated along coreference chains and the original sequence in parallel. 

We compare our Coref-GRU layer with the regular GRU layer by incorporating it in a recent model for reading comprehension. On synthetic data specifically constructed to test coreference-based reasoning \citep{weston2015towards}, C-GRUs lead to a large improvement over regular GRUs. We show that the structural bias introduced and coreference signals are both important to reach high performance in this case. On a more realistic dataset \citep{welbl2017constructing}, with noisy coreference annotations, we see small but significant improvements over a state-of-the-art baseline. As we reduce the training data, the gains become larger.
Lastly, we apply the same model to a broad-context language modeling task \citep{paperno2016lambada}, where coreference resolution is an important factor, and show improved performance over state-of-the-art.

\section{Related Work}

\textbf{Entity-based models.} \citet{ji2017dynamic} presented a generative model for jointly predicting the next word in the text and its gold-standard coreference annotation. The difference in our work is that we look at the task of reading comprehension, and also work in the more practical setting of system extracted coreferences. EntNets \citep{henaff2016tracking} also maintain dynamic memory slots for entities, but do not use coreference signals and instead update all memories after reading each sentence, which leads to poor performance in the low-data regime (c.f. Table \ref{tab:babi}).
\citet{yang2016reference} model references in text as explicit latent variables, but limit their work to text generation. \citet{kobayashi2016dynamic} used a pooling operation to aggregate entity information across multiple mentions.
\citet{wang2016emergent} also noted the importance of reference resolution for reading comprehension, and we compare our model to their one-hot pointer reader.

\textbf{Syntactic-recency.} Recent work has used syntax, in the form of dependency trees, to replace the sequential recency bias in RNNs with a syntactic recency bias \citep{tai2015improved,swayamdipta2017learning,qian2017syntax,chen2017improved}. However, syntax only looks at dependencies within sentence boundaries, whereas our focus here is on longer ranges. Our resulting layer is structurally similar to GraphLSTMs \citep{peng2017cross}, with an additional attention mechanism over the graph edges. However, while \citet{peng2017cross} found that using coreference did not lead to any gains for the task of relation extraction, here we show that it has a positive impact on the reading comprehension task.
\textbf{Self-Attention} \citep{vaswani2017attention} models are becoming popular for modeling long-term dependencies, and may also benefit from coreference information to bias the learning of those dependencies. Here we focus on recurrent layers and leave such an analysis to future work.

Part of this work was described in an unpublished preprint \citep{dhingra2017linguistic}. The current paper extends that version and focuses exclusively on coreference relations. We also report results on the WikiHop dataset, including the performance of the model in the low-data regime.

\section{Model}

\begin{figure}
	\centering
    \includegraphics[width=0.8\linewidth]{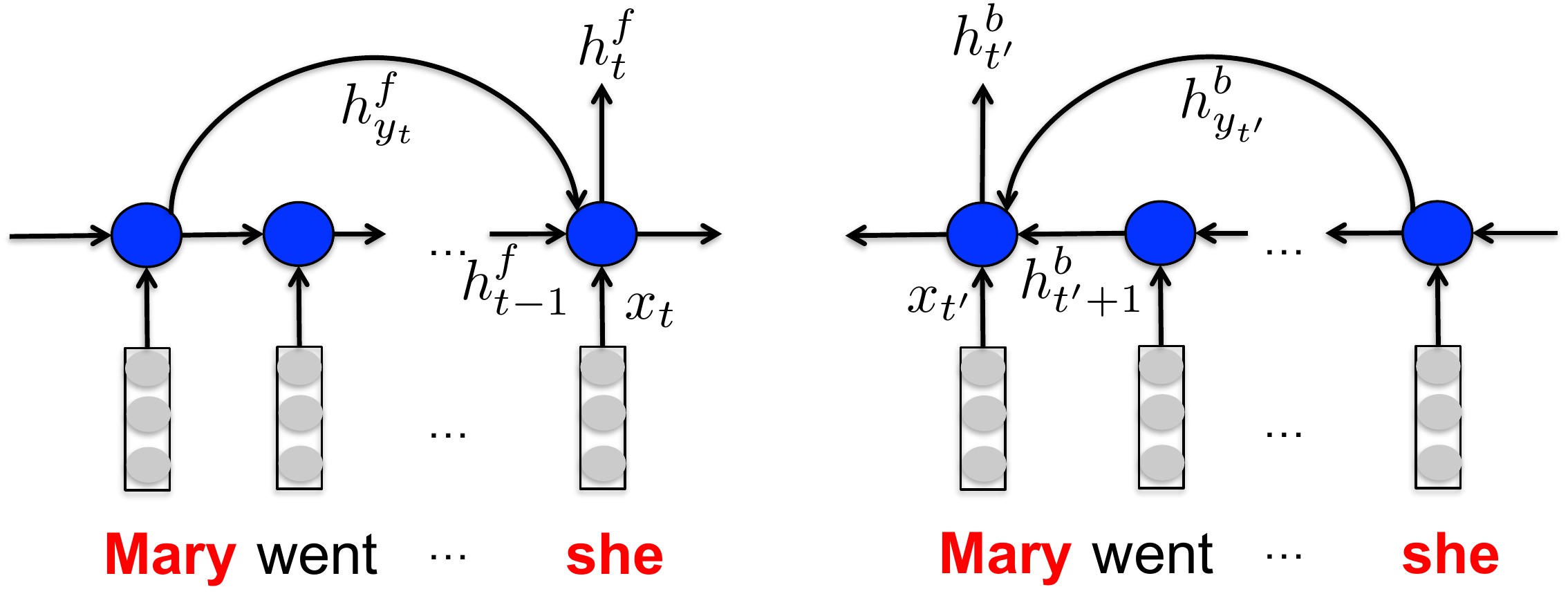}
    \caption{\small Forward (left) and Backward (right) Coref-GRU layers. \textbf{Mary} and \textbf{she} are coreferent.}\label{fig:coref-gru}
\end{figure}

\textbf{Coref-GRU (C-GRU) Layer.} Suppose we are given an input sequence $w_1, w_2, \ldots, w_T$ along with their word vectors $x_1, \ldots, x_T$ and annotations for the most recent coreferent antecedent for each token $y_1, \ldots, y_T$, where $y_t \in \{0, \ldots, t-1\}$ and $y_t=0$ denotes the null antecedent (for tokens not belonging to any cluster). We assume all tokens belonging to a mention in a cluster belong to that cluster, and there are $C$ clusters in total. 
Our recurrent layer is adapted from GRU cells \citep{cho2014learning}, but similar extensions can be derived for other recurrent cells as well. The update equations in a GRU all take the same basic form given by:
\begin{equation*}
    f(Wx_t + Uh_{t-1} + b).
\end{equation*}
The bias for sequential recency comes from the second term $Uh_{t-1}$. In this work we add another term to introduce a bias towards coreferent recency instead:
\begin{equation*}
    f(Wx_t + \alpha_t U \phi_s(h_{t-1}) + (1-\alpha_t) U' \phi_c(h_{y_t}) + b),
\end{equation*}
where $h_{y_t}$ is the hidden state of the coreferent antecedent of $w_t$ (with $h_0=0$), $\phi_s$ and $\phi_c$ are non-linear functions applied to the hidden states coming from the sequential antecedent and the coreferent antecedent, respectively, and
$\alpha_t$ is a scalar weight which decides the relative importance of the two terms based on the current input (so that, for example, pronouns may assign a higher weight for the coreference state). 
When $y_t = 0$, $\alpha_t$ is set to 1, otherwise it is computed using a key-based addressing scheme \citep{miller2016key}, as $\alpha_t = \text{softmax}(x_t^T k)$, where $k$ is a trainable key vector. 
In this work we use simple slicing functions $\phi_s(x) = x[1:d/2]$, and $\phi_c(x)=x[d/2:d]$ which decompose the hidden states into a sequential and a coreferent component, respectively. Figure \ref{fig:coref-gru} (left) shows an illustration of the layer, and the full update equations are given in Appendix \ref{app:equations}.

\textbf{Connection to Memory Networks.} We can also view the model as a memory network \citep{sukhbaatar2015end} with a memory state $M_t$ at each time step which is a $C\times d$ matrix. The rows of this memory matrix correspond to the state of each coreference cluster at time step $t$. The main difference between Coref-GRUs and a typical memory network such as EntNets \citep{henaff2016tracking} lies in the fact that we use coreference annotations to read and write from the memory rather than let model learn how to access the memory. With Coref-GRUs, only the content of the memories needs to be learned. As we shall see in Section \ref{sec:experiments}, this turns out to be a useful bias in the low-data regime.

\textbf{Bidirectional C-GRU.} To extend to the bidirectional case, a second layer is fed the same sequence in the reverse direction, $x_T, \ldots, x_1$ and $y_t \in \{0, t+1, \ldots, T\}$ now denotes the immediately descendent coreferent token from $w_t$. Outputs from the two layers are then concatenated to form the bi-directional output (see Figure \ref{fig:coref-gru}). 

\textbf{Complexity.} The resulting layer has the same time-complexity as that of a regular GRU layer. The memory complexity increases since we have to keep track of the hidden states for each coreference cluster in the input. If there are $C$ clusters and $B$ is the batch size, the resulting complexity is by $O(BTCd)$. This scales linearly with the input size $T$, however we leave exploration of more efficient architectures to future work.


\textbf{Reading comprehension architecture.} All tasks we look at involve tuples of the form $(p, q, a, \mathcal{C})$, where the goal is to find the answer $a$ from candidates $\mathcal{C}$ to question $q$ with passage $p$ as context. We use the Gated-Attention (GA) reader \citep{dhingra2016gated} as a base architecture, which computes representations of the passage by passing it through multiple bidirectional GRU layers with an attention mechanism in between layers.
We compare the original GA architecture (GA w/ GRU) with one where the bidirectional GRU layers are replaced with bidirectional C-GRU layers (GA w/ C-GRU). Performance is reported in terms of the accuracy of detecting the correct answer from $\mathcal{C}$, and all models are trained using cross-entropy loss. When comparing two models we ensure the number of parameters are the same in each. Other implementation details are listed in Appendix \ref{app:hyperparams}.

\section{Experiments \& Results}
\label{sec:experiments}


\begin{table}[!htbp]
\centering
\small
\begin{tabular}{@{}lccc@{}}
\toprule
\textbf{Method}      & \textbf{Avg} & \textbf{Max}   & \textbf{\# failed} \\ \midrule
EntNets \citep{henaff2016tracking}              & --           & 0.704          & 15                 \\
QRN \citep{seo2016query}                  & --           & \textbf{0.901} & 7                  \\ \midrule
Bi-GRU                & 0.727        & 0.767          & 13                 \\
Bi-C-GRU             & 0.790        & 0.831          & 12                 \\
GA w/ GRU                   & 0.764        & 0.810          & 10                 \\
GA w/ GRU + 1-hot         & 0.766        & 0.808          & 9                  \\
GA w/ C-GRU    & 0.870        & 0.886          & \textbf{5}         \\ \bottomrule
\end{tabular}
\caption{Accuracy on bAbi-1K, averaged across all $20$ tasks. Following previous work we run each task for $10$ random seeds, and report the \textbf{Avg} and \textbf{Max} (based on dev set) performance. A task is considered failed if its \textbf{Max} performance is $<$ $0.95$.}

\label{tab:babi}
\end{table}

\textbf{BAbi AI tasks.} Our first set of experiments are on the 1K training version of the synthetic bAbi AI tasks \citep{weston2015towards}. The passages and questions in this dataset are generated using templates, removing many complexities inherent in natural language, but it still provides a useful testbed for us since some tasks are specifically constructed to test the coreference-based reasoning we tackle here. Experiments on more natural data are described below.

Table \ref{tab:babi} shows a comparison of EntNets \citep{henaff2016tracking}, QRNs \citep{seo2016query} (the best published results on bAbi-1K), and our models. We also include the results for a single layer version of GA Reader (which we denote simply as Bi-GRU or Bi-C-GRU when using coreference) to enable fair comparison with EntNets. In each case we see clear improvements of using C-GRU layers over GRU layers. Interestingly, EntNets, which have $>$$99\%$ performance when trained with $10K$ examples only reach $70\%$ performance with 1K training examples. The Bi-C-GRU model significantly improves on this baseline, which shows that, with less data, coreference annotations can provide a useful bias for a memory network on how to read and write memories.

A break-down of task-wise performance is given in Appendix \ref{app:babi-tasks}. Comparing C-GRU to the GRU based method, we find that the main gains are on tasks 2 (two supporting facts), 3 (three supporting facts) and 16 (basic induction). All these tasks require aggregation of information across sentences to derive the answer. Comparing to the QRN baseline, we found that C-GRU was significantly worse on task 15 (basic deduction). On closer examination we found that this was because our simplistic coreference module which matches tokens exactly was not able to resolve ``mice'' to ``mouses'' and ``cats'' to ``cat''. On the other hand, C-GRU was significantly better than QRN on task 16 (basic induction).

\begin{figure}
	\centering
    \includegraphics[width=0.48\linewidth, trim={0mm 0mm 12mm 0mm}, clip]{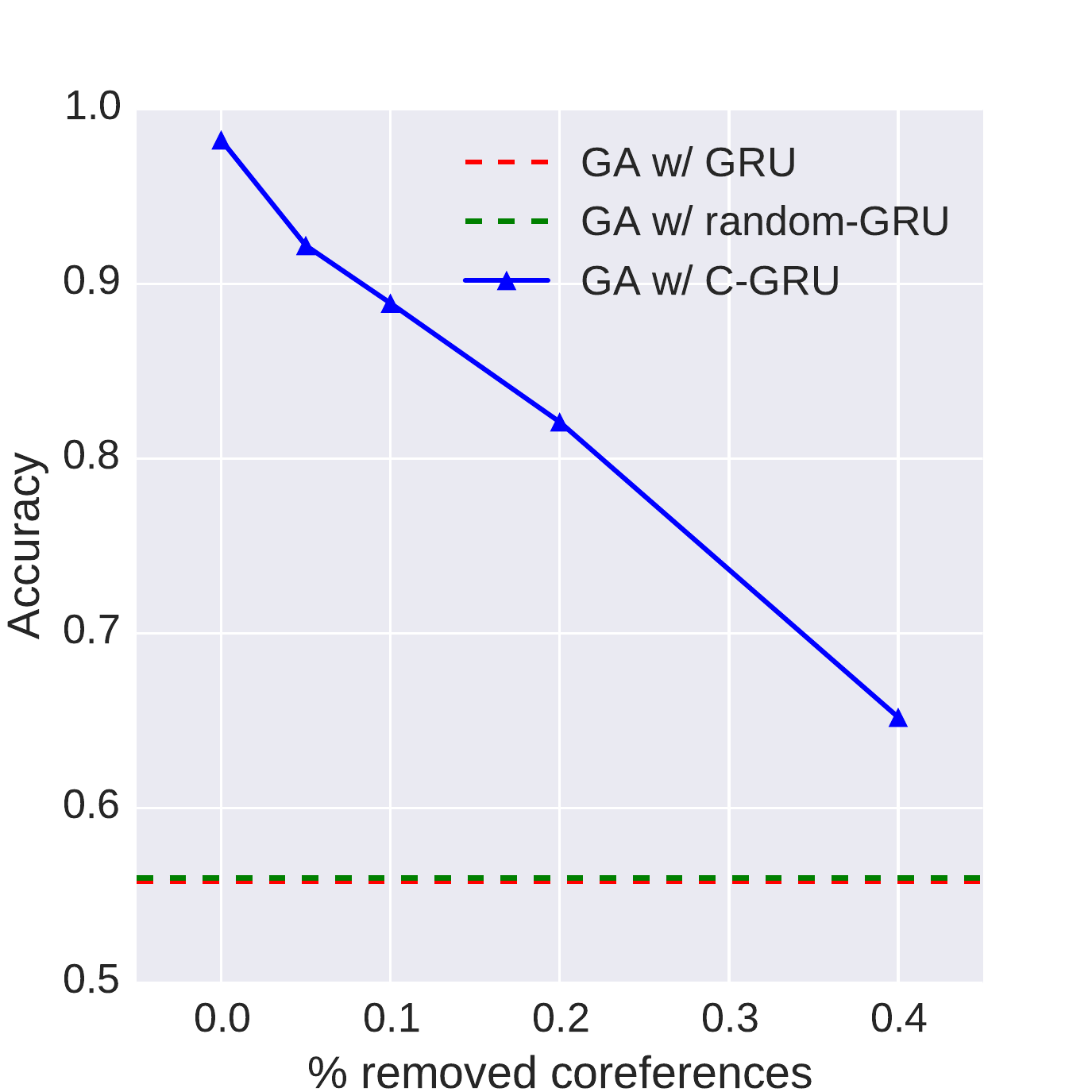}~
    \includegraphics[width=0.48\linewidth, trim={0mm 0mm 12mm 0mm}, clip]{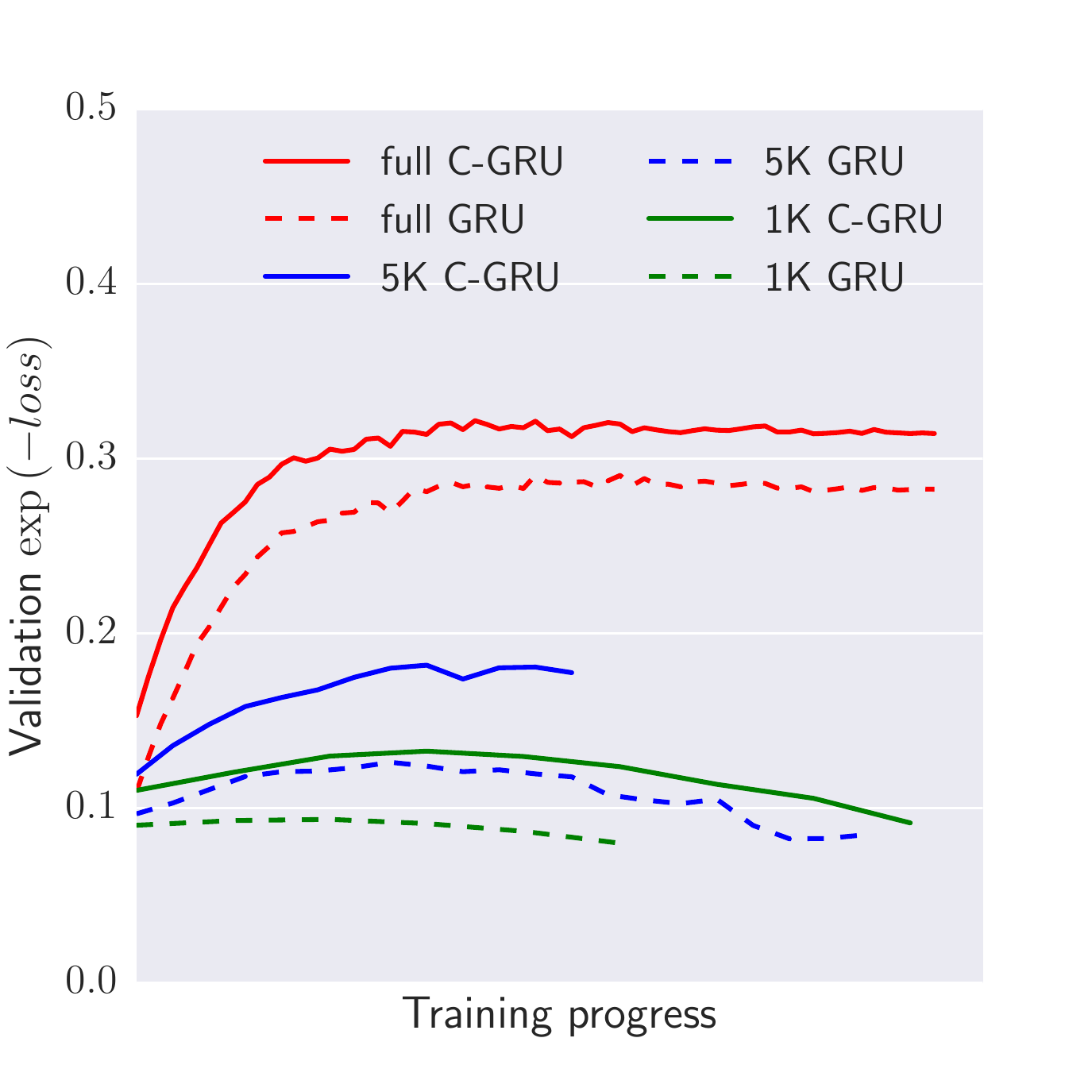}
    \caption{\small \textbf{Left:} Accuracy of GA w/ C-GRU as coreference annotations are removed for bAbi task 3. \textbf{Right:} Expected probability of correct answer ($\exp{(-loss)}$) on Validation set as training progresses on Wikihop dataset for 1K, 5K and the full training datasets.}\label{fig:noise}
\end{figure}

We also include a baseline which uses coreference features as 1-hot vectors appended to the input word vectors (GA w/ GRU + 1-hot). This provides the model with information about the coreference clusters, but does not improve performance, suggesting that the regular GRU is unable to track the given coreference information across long distances to solve the task. On the other hand, in Figure \ref{fig:noise} (left) we show how the performance of GA w/ C-GRU varies as we remove gold-standard  mentions from coreference clusters, or if we replace them with random mentions (GA w/ random-GRU). In both cases there is a sharp drop in performance, showing that specifically using coreference for connecting mentions is important.

\begin{table}[!htbp]
\setlength\tabcolsep{2.75pt}
\centering
\small
\begin{tabular}{@{}lccccc@{}}
\toprule
\multirow{2}{*}{\textbf{Method}} & \textbf{Follow} & \textbf{\begin{tabular}[c]{@{}c@{}}Follow\\ +single\end{tabular}} & \textbf{\begin{tabular}[c]{@{}c@{}}Follow\\ +multiple\end{tabular}} & \multicolumn{2}{c}{\textbf{Overall}} \\ \cmidrule(l){2-6} 
                                 & \textbf{Dev}     & \textbf{Dev}                                                        & \textbf{Dev}                                                          & \textbf{Dev}      & \textbf{Test}    \\ \midrule
\multicolumn{6}{l}{\textbf{1K}}                                                                                                                                                                                                          \\ \midrule
GA w/ GRU                        & 0.307           & 0.332                                                              & 0.287                                                                & 0.263            & --               \\
GA w/ C-GRU                      & 0.355           & 0.370                                                              & 0.354                                                                & 0.330            & --               \\ \midrule
\multicolumn{6}{l}{\textbf{5K}}                                                                                                                                                                                                          \\ \midrule
GA w/ GRU                        & 0.382           & 0.385                                                              & 0.390                                                                & 0.336            & --               \\
GA w/ C-GRU                      & 0.452           & 0.454                                                              & 0.460                                                                & 0.401            & --               \\ \midrule
\multicolumn{6}{l}{\textbf{full}}                                                                                                                                                                                                        \\ \midrule
BiDAF                            & --               & --                                                                  & --                                                                    & --                & 0.429           \\
GA w/ GRU                        & 0.606           & 0.615                                                              & 0.604                                                                & 0.549            & --               \\
GA w/ C-GRU                      & \textbf{0.614}  & \textbf{0.616}                                                     & \textbf{0.614}                                                       & \textbf{0.560}$^\dagger$   & \textbf{0.593}  \\ \bottomrule
\end{tabular}
\caption{Accuracy on Wikihop. \textbf{Follow:} annotated as answer follows from the given passages. \textbf{Follow +multiple:} annotated as requiring multiple passages for answering. \textbf{Follow +single} annotated as requiring one passage for answering.
$^\dagger p=0.057$ using Mcnemar's test compared to GA w/ GRU.}
\label{tab:wikihop}
\end{table}


\textbf{Wikihop dataset.} Next we apply our model to the Wikihop dataset \citep{welbl2017constructing}, which is specifically constructed to test multi-hop reading comprehension across documents. Each instance in this dataset consists of a collection of passages $(p_1, \ldots, p_N)$, and a query of the form $(h, r)$ where $h$ is an entity and $r$ is a relation. The task is to find the tail entity $t$ from a set of provided candidates $\mathcal{C}$. As preprocessing we concatenate all documents in a random order, and extract coreference annotations from the Berkeley Entity Resolution system \citep{durrett2013easy} which gets about $62\%$ F1 score on the CoNLL 2011 test set.
We only keep the coreference clusters which contain at least one candidate from $\mathcal{C}$ or an entity which co-occurs with the head entity $h$. We report results in Table \ref{tab:wikihop} when using the full training set, as well as when using a reduced training set of sizes 1K and 5K, to test the model under a low-data regime. In Figure \ref{fig:noise} we also show the training curves of $\exp{(-loss)}$ on the validation set.

We see higher performance for the C-GRU model in the low data regime, and better generalization throughout the training curve for all three settings. This supports our conjecture that the GRU layer has difficulty learning the kind of coreference-based reasoning required in this dataset, and that the bias towards coreferent recency helps with that. However, perhaps surprisingly, given enough data both models perform comparably. This could either indicate that the baseline learns the required reasoning patterns when given enough data, or, that the bias towards corefence-based reasoning hurts performance for some other types of questions. Indeed, there are $9\%$ questions which are answered correctly by the baseline but not by C-GRU, however, we did not find any consistent patterns among these in our analyses. Lastly, we note that both models vastly outperform the best reported result of BiDAf from \citep{welbl2017constructing}\footnote{The official leaderboard at \url{http://qangaroo.cs.ucl.ac.uk/leaderboard.html} shows two models with better performance than reported here (as of April 2018). Since we were unable to find publications for these models we omit them here.}. We believe this is because the GA models select answers from the list of candidatees, whereas BiDAF ignores those candidates.

\begin{table}[!htbp]
\small
\centering
\begin{tabular}{@{}lcc@{}}
\toprule
\textbf{Method}   & \textbf{overall} & \textbf{context} \\ \midrule
\citet{chu2016broad} & 0.4900               & --           \\
GA w/ GRU         &  0.5398          &   0.6677         \\
GA w/ GRU + 1-hot &     0.5338        &   0.6603        \\
GA w/ C-GRU       &   \textbf{0.5569}         &  \textbf{0.6888}$^\dagger$          \\ \bottomrule
\end{tabular}
\caption{Accuracy on LAMBADA test set, averaged across two runs with random initializations. \textbf{context:} passages for which the answer is in context. \textbf{overall:} full test set for comparison to prior work. $^\dagger p < 0.0001$ using Mcnemar's test compared to GA w/ GRU.}
\label{tab:lambada}
\end{table}

\textbf{LAMBADA dataset.} Our last set of experiments is on the broad-context language modeling task of LAMBADA dataset \citep{paperno2016lambada}. This dataset consists of passages 4-5 sentences long, where the last word needs to be predicted. Interestingly, though, the passages are filtered such that human volunteers were able to predict the missing token given the full passage, but not given only the last sentence. Hence, predicting these tokens involves a broader understanding of the whole passage. Analysis of the questions \citep{chu2016broad} suggests that around 20\% of the questions need coreference understanding to answer correctly. Hence, we apply our model which uses coreference information for this task.

We use the same setup as \citet{chu2016broad} which formulated the problem as a reading comprehension one by treating the last sentence as query, and the remaining passage as context to extract the answer from. In this manner only $80\%$ of the questions are answerable, but the performance increases substantially compared to pure language modeling based approaches. For this dataset we used Stanford CoreNLP to extract coreferences \citep{clark2015entity}, which achieved $0.63$ F1 on the CoNLL test set. Table \ref{tab:lambada} shows a comparison of the GA w/ GRU baseline and GA w/ C-GRU models. We see a significant gain in performance when using the layer with coreference bias. Furthermore, the 1-hot baseline which uses the same coreference information, but with sequential recency bias fails to improve over the regular GRU layer. While the improvement for C-GRU is small, it is significant, and we note that questions in this dataset involve several different types of reasoning out of which we only tackle one specific kind. The proposed GA w/ C-GRU layer sets a new state-of-the-art on this dataset.

\section{Conclusion}

We present a recurrent layer with a bias towards \textit{coreferent recency}, with the goal of tackling reading comprehension problems which require aggregating information from multiple mentions of the same entity. Our experiments show that when combined with a powerful reading architecture, the layer provides a useful inductive bias for solving problems of this kind. In future work, we aim to apply this model to other problems where long-term dependencies at the document level are important. Noise in the coreference annotations has a detrimental effect on the performance (Figure \ref{fig:noise}), hence we also aim to explore joint models which learn to do coreference resolution and reading together.

\section*{Acknowledgments}
This work was supported by NSF under grants CCF-1414030 and IIS-1250956 and by grants from Google.

\bibliography{naaclhlt2018}
\bibliographystyle{acl_natbib}

\appendix

\section{C-GRU update equations}
\label{app:equations}

For simplicity, we introduce the variable $m_t$ which concatenates ($||$) the sequential and coreferent hidden states:
\begin{align*}
    m_t &= \alpha_t \phi_s(h_{t-1}) || (1-\alpha_t) \phi_c(h_{y_t})
\end{align*}

\noindent Then the update equations are given by:
\begin{align*}
    r_t &= \sigma(W^r x_t + U^r m_t + b^r)\\
    z_t &= \sigma(W^z x_t + U^zm_t + b^z)\\
    \tilde{h}_t &= \tanh(W^h x_t + r_t \odot U^hm_t + b^h)\\
    h_t &= (1-z_t) \odot m_t + z_t \tilde{h}_t
\end{align*}

\noindent The attention parameter $\alpha_t$ is given by:
\begin{equation*}
    \alpha_t = \frac{\exp{x_t^T k_1}}{\exp{x_t^T k_1} + \exp{x_t^T k_2}}
\end{equation*}
where $k_1$ and $k_2$ are trainable key vectors.

\section{Implementation details}
\label{app:hyperparams}

We use $K=3$ layers with the GA architecture. We keep the same hyperparameter settings when using GRU or C-GRU layers, which we describe here.

For the \textbf{bAbi} dataset, we use a hidden state size of $d=64$, batch size of $B=32$, and learning rate $0.01$ which is halved after every $120$ updates. We also use dropout with rate $0.1$ at the output of each layer. The maximum number of coreference clusters across all tasks was $C=13$. Half of the tasks in this dataset are extractive, meaning the answer is present in the passage, whereas the other half are classification tasks, where the answer is in a list of candidates which may not be in the passage. For the extractive tasks, we use the attention sum layer as described in the GA Reader paper \citep{dhingra2016gated}. For the classification tasks we replace this with a softmax layer for predicting one of the classes.

For the \textbf{Wikihop} dataset, we use a hidden state size of $d=64$, batch size $B=16$, and learning rate of $0.0005$ which was halved every $2500$ updates. The maximum number of coreference clusters was set to $50$ for this dataset. We used dropout of $0.2$ in between the intermediate layers, and initialized word embeddings with Glove \citep{pennington2014glove}. We also used character embeddings, which were concatenated with the word embeddings, of size $10$. These were output from a CNN layer with $50$ filters each of width $5$. Following \citep{weissenborn2017fastqa}, we also appended a feature to the word embeddings in the passage which indicated if the token appeared in the query or not. 

For the \textbf{LAMBADA} dataset, we use a hidden state size of $d=256$, batch size of $B=64$, and learning rate of $0.0005$ which was halved every $2$ epochs. Word vectors were initialized with Glove, and dropout of $0.2$ was applied after intermediate layers. The maximum number of coreference clusters in this dataset was $15$.

\section{Task-wise bAbi performance}
\label{app:babi-tasks}

\begin{table}[!htbp]
\small
\centering
\begin{tabular}{@{}lccc@{}}
\toprule
\textbf{Task}                      & \textbf{QRN}   & \textbf{\begin{tabular}[c]{@{}c@{}}GA w/\\ GRU\end{tabular}} & \textbf{\begin{tabular}[c]{@{}c@{}}GA w/\\ C-GRU\end{tabular}} \\ \midrule
1: Single Supporting Fact          & 1.000          & 0.997                                                        & 1.000                                                          \\
\textbf{2: Two Supporting Facts}   & \textbf{0.993} & \textbf{0.345}                                               & \textbf{0.990}                                                 \\
\textbf{3: Three Supporting Facts} & \textbf{0.943} & \textbf{0.558}                                               & \textbf{0.982}                                                 \\
4: Two Argument Relations          & 1.000          & 1.000                                                        & 1.000                                                          \\
5: Three Argument Relations        & 0.989          & 0.989                                                        & 0.993                                                          \\
6:Yes/No Questions                 & 0.991          & 0.962                                                        & 0.976                                                          \\
\textbf{7: Counting}              & \textbf{0.904} & \textbf{0.946}                                               & \textbf{0.976}                                                 \\
8: Lists / Sets                    & 0.944          & 0.947                                                        & 0.964                                                          \\
9: Simple Negation                 & 1.000          & 0.991                                                        & 0.990                                                          \\
10: Indefinite Knowledge           & 1.000          & 0.992                                                        & 0.986                                                          \\
11: Basic Coreference              & 1.000          & 0.995                                                        & 0.996                                                          \\
12: Conjunction                    & 1.000          & 1.000                                                        & 0.996                                                          \\
13: Compound Coreference           & 1.000          & 0.998                                                        & 0.993                                                          \\
\textbf{14: Time Reasoning}        & \textbf{0.992} & \textbf{0.895}                                               & \textbf{0.849}                                                 \\
\textbf{15: Basic Deduction}       & \textbf{1.000} & \textbf{0.521}                                               & \textbf{0.470}                                                 \\
\textbf{16: Basic Induction}       & \textbf{0.470} & \textbf{0.488}                                               & \textbf{0.999}                                                 \\
17: Positional Reasoning           & 0.656          & 0.580                                                        & 0.574                                                          \\
18: Size Reasoning                 & 0.921          & 0.908                                                        & 0.896                                                          \\
19: Path Finding                   & 0.213          & 0.095                                                        & 0.099                                                          \\
20: Agent's Motivation             & 0.998          & 0.998                                                        & 1.000                                                          \\ \midrule
\textbf{Average}                   & \textbf{0.901} & \textbf{0.810}                                               & \textbf{0.886}                                                 \\ \bottomrule
\end{tabular}
\caption{Breakdown of task-wise performance on bAbi dataset. Tasks where C-GRU is significant better / worse than either GRU or QRNs are highlighted.}
\label{my-label}
\end{table}

\end{document}